\documentclass[letterpaper]{article} 
\usepackage{aaai25}  
\usepackage{times}  
\usepackage{helvet}  
\usepackage{courier}  
\usepackage[hyphens]{url}  
\usepackage{graphicx} 
\usepackage{natbib}  
\usepackage{caption} 
\usepackage{algorithm}
\usepackage{algorithmic}

\usepackage{newfloat}
\usepackage{listings}
\DeclareCaptionStyle{ruled}{labelfont=normalfont,labelsep=colon,strut=off} 
\floatstyle{ruled}
\newfloat{listing}{tb}{lst}{}
\floatname{listing}{Listing}
\usepackage{microtype}
\usepackage{inconsolata}
\usepackage{multirow}
\usepackage{makecell}
\usepackage{enumitem}
\usepackage{amsmath, amssymb}
\usepackage{diagbox}
\usepackage{booktabs}
\usepackage{multirow}
\usepackage{bbding}
\renewcommand{\algorithmiccomment}[1]{\hfill// #1}
\usepackage{dsfont}
\usepackage{tabularx}

\title{\textsc{ReFF}: Reinforcing Format Faithfulness in Language Models across Varied Tasks}
\author{
    Jiashu Yao\textsuperscript{\rm 1},
    Heyan Huang\textsuperscript{\rm 1},
    Zeming Liu\textsuperscript{\rm 2},
    Haoyu Wen\textsuperscript{\rm 1},
    Wei Su\textsuperscript{\rm 1},
    Boao Qian\textsuperscript{\rm 1},
    Yuhang Guo\textsuperscript{\rm 1}\footnote{Corresponding author}
}
\affiliations{
    \textsuperscript{\rm 1}School of Computer Science and Technology, Beijing Institute of Technology\\
    \textsuperscript{\rm 2}School of Computer Science and Engineering, Beihang University\\
    \texttt{\{yaojiashu, hhy63, haoyuwen, weisu, qianboao, guoyuhang\}@bit.edu.cn, zmliu@buaa.edu.cn}
}

\begin{document}

\maketitle

\begin{abstract}
Following formatting instructions to generate well-structured content is a fundamental yet often unmet capability for large language models (LLMs). To study this capability, which we refer to as format faithfulness, we present \textsc{FormatBench}, a comprehensive format-related benchmark. Compared to previous format-related benchmarks, \textsc{FormatBench} involves a greater variety of tasks in terms of application scenes (traditional NLP tasks, creative works, autonomous agency tasks), human-LLM interaction styles (single-turn instruction, multi-turn chat), and format types (inclusion, wrapping, length, coding). Moreover, each task in \textsc{FormatBench} is attached with a format checker program. Extensive experiments on the benchmark reveal that state-of-the-art open- and closed-source LLMs still suffer from severe deficiency in format faithfulness. By virtue of the decidable nature of formats, we propose to Reinforce Format Faithfulness (\textsc{ReFF}) to help LLMs generate formatted output as instructed without compromising general quality. Without any annotated data, \textsc{ReFF} can substantially improve the format faithfulness rate (e.g., from 21.6\% in original LLaMA3 to 95.0\% on caption segmentation task), while keep the general quality comparable (e.g., from 47.3 to 46.4 in F1 scores). Combined with labeled training data, \textsc{ReFF} can simultaneously improve both format faithfulness (e.g., from 21.6\% in original LLaMA3 to 75.5\%) and general quality (e.g., from 47.3 to 61.6 in F1 scores). We further offer an interpretability analysis to explain how \textsc{ReFF} improves both format faithfulness and general quality.
\end{abstract}

%
\begin{links}
    \link{Code \& Datasets}{https://github.com/BITHLP/ReFF}
\end{links}

\section{Introduction}

Recent years have witnessed a significant upsurge in the development and deployment of large language models (LLMs) \cite{brown2020language, touvron2023llama2, achiam2023gpt}. With their exceptional zero-shot and few-shot capabilities, LLMs have revolutionized the paradigm of language-related tasks, where a question can be understood and solved to the best without task-specific supervision \cite{radford2019language}.

\begin{figure}[htb!]
    \centering
    \includegraphics[width=0.9\linewidth]{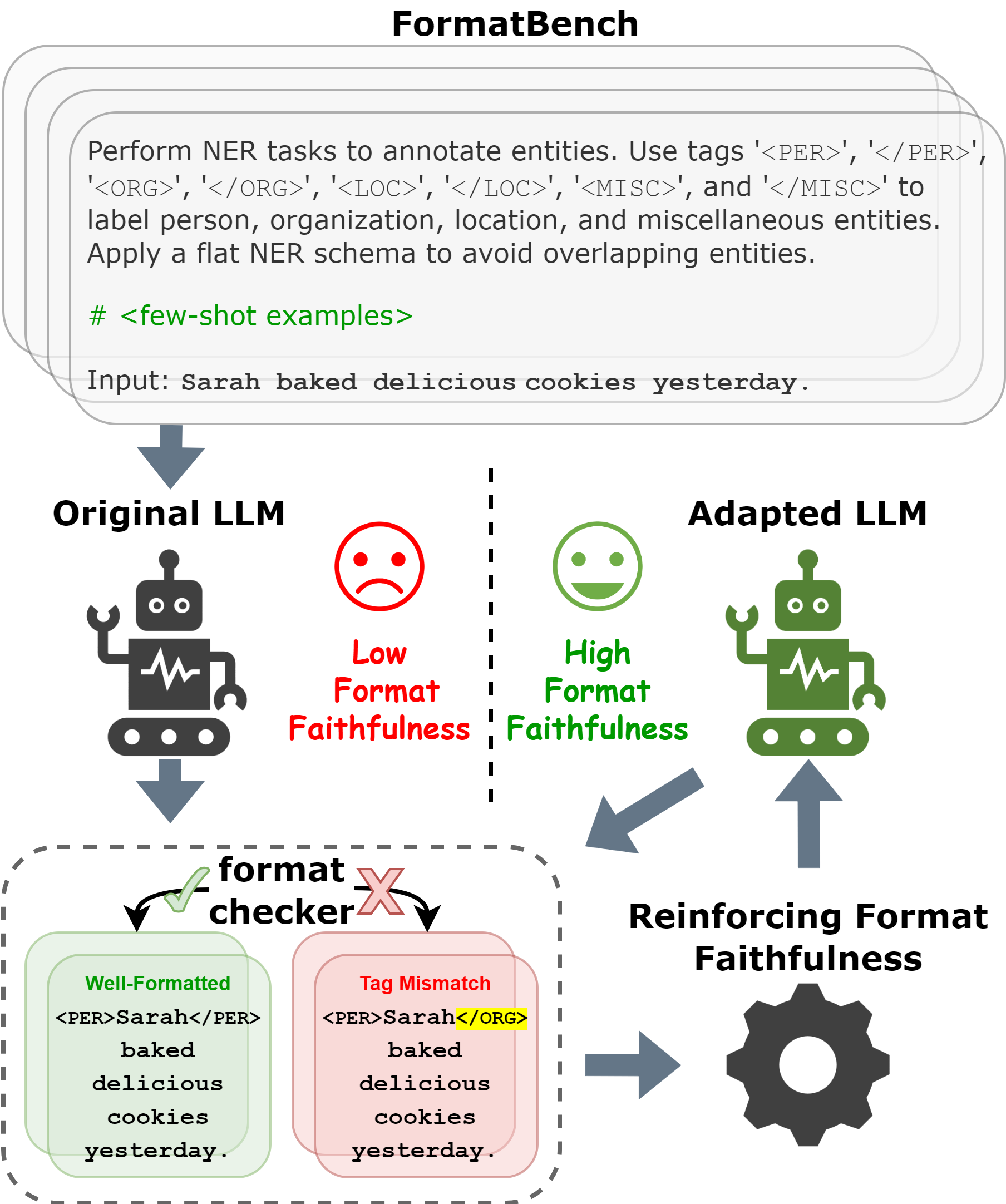}
    \caption{The overall framework of this work. The queries in \textsc{FormatBench} are forwarded to an LLM to generate corresponding responses, whose format correctness are labelled by a format checker. The queries, generated responses, and the format labels are utilized in \textsc{ReFF} process to iteratively obtain an adapted LLM with higher format faithfulness.}
    \label{fig:reff}
\end{figure}

The zero- and few-shot prompting paradigms have introduced a new problem in task solving procedure, namely, the specification of the output format. To elaborate, LLMs' task solving paradigm mandates that users must devise an output format and include it in the prompt as a request for LLMs to adhere to. The format specification holds significant importance in tasks of wide concern, for example:

\begin{table*}[htbp]
    \centering
    \begin{tabular}{ccccccccccc} 
    \toprule
        \multirow{2}{*}{\textbf{Name}} & \multirow{2}{*}{\textbf{\# Test}} & \multicolumn{3}{c}{\textbf{Application Scene}} & \multicolumn{2}{c}{\textbf{Interaction Style}} & \multicolumn{4}{c}{\textbf{Format Type}} \\

        \cmidrule(lr){3 - 5}
        \cmidrule(lr){6 - 7}
        \cmidrule(lr){8 - 11}

        & & \textbf{TradNLP} & \textbf{Creat} & \textbf{Robot} & \textbf{Single} & \textbf{Multi} & \textbf{Include} & \textbf{Wrap} & \textbf{Length} & \textbf{Code} \\
        \hline

        \textsc{Chem-*} & 148 & \XSolidBrush & \XSolidBrush & \CheckmarkBold & \CheckmarkBold & \XSolidBrush & \XSolidBrush & \XSolidBrush & \XSolidBrush & \CheckmarkBold \\
        \textsc{S-Bench} & 1,727 & \XSolidBrush & \CheckmarkBold & \XSolidBrush & \CheckmarkBold & \XSolidBrush & \CheckmarkBold & \XSolidBrush & \XSolidBrush & \CheckmarkBold \\
        \textsc{IFEval} & 541 & \XSolidBrush & \CheckmarkBold & \XSolidBrush & \CheckmarkBold & \XSolidBrush & \CheckmarkBold & \CheckmarkBold & \CheckmarkBold & \XSolidBrush \\
        \textsc{FoFo} & 494 & \XSolidBrush & \CheckmarkBold & \XSolidBrush & \CheckmarkBold & \XSolidBrush & \CheckmarkBold & \CheckmarkBold & \XSolidBrush & \CheckmarkBold \\
        \hline
        \textsc{FormatBench} & 24,483 & \CheckmarkBold & \CheckmarkBold & \CheckmarkBold & \CheckmarkBold & \CheckmarkBold & \CheckmarkBold & \CheckmarkBold & \CheckmarkBold & \CheckmarkBold \\
        \bottomrule
    \end{tabular}
    \caption{Comparison between \textsc{FormatBench} and previous format-related benchmarks. \textsc{FormatBench} features a significantly larger test set, offering a greater variety of application scenes, human-computer interaction styles, and format requirement types. The definition of each category is described in detail in Section \ref{subsec:variety}.}
    \label{tab:bench-compare}
\end{table*}

\begin{itemize}
    \item Various natural language processing (NLP) tasks, such as named entity recognition, text-to-data conversion, and syntactic parsing, hold rigorous format requirements.
    \item Creative works, such as poems, intrinsically possess rigorous forms, including acrostic and numerous others.
    \item LLM-based autonomous agents need strict format adherence to avoid system crashes or dangerous behaviors.
\end{itemize}

To summarize, the ability to adhere to pre-defined format specifications is of utmost importance in the deployment of LLMs. This ability, which we refer to as format faithfulness, is a crucial aspect to consider in many real-world tasks.

However, there still exists two significant gaps in studies relating to format faithfulness. Firstly, current datasets related to formatting are primarily focused on one specific task, such as text-to-data \cite{tang2023struc}, code generation \cite{skreta2023errors}, one-turn instruction \cite{li2024instruction}, and specialty area documentation \cite{xia2024fofo}, rather than covering varied tasks. This narrow focus restricts the breadth and reliability of format faithfulness evaluation. Secondly, current adaptation approaches aimed at improving format faithfulness like prompt engineering \cite{skreta2023errors} and finetuning \cite{tang2023struc} neglect the decidable nature of format problems, i.e., whether a response adheres to format requirements can be assessed by a non-parameter format checker. This oversight can result in lower effectiveness, as demonstrated in the following experiments.

To address the gap in comprehensive benchmarks, we combine adaptation of existing datasets, online data collection, and manual data annotation, presenting \textsc{FormatBench}. Compared to previous benchmarks, \textsc{FormatBench} includes not only a significantly larger test set, but also a wider range of tasks in diverse application scenes, interaction styles, and format types. As a result, it obtains a comprehensive evaluation of the format faithfulness of LLMs. Extensive experiments on the benchmark reveal that \textsc{FormatBench} poses significant challenges to even the most capable models with simple format requirements, such as selecting among admissible options in a multi-choice question.

To fill the gap in format adaptation approaches of neglecting format decidability, we propose Reinforcing Format Faithfulness (\textsc{ReFF}), as is illustrated in Figure \ref{fig:reff}. \textsc{ReFF} takes full advantage of the decidability of format by using a format checker to judge the format correctness of LLM generated content, and then utilizing the judged data in a reinforcement learning (RL) process to improve format faithfulness. Extensive experiments of \textsc{ReFF} on \textsc{FormatBench} yield highly favorable results. Without any annotated data, \textsc{ReFF} can significantly improve the format faithfulness rate (e.g., from 21.6\% in original LLaMA3 to 95.0\% on caption segmentation task), while keep the general quality comparable (e.g., from 47.3 to 46.4 in F1 scores). Combined with labeled training data, \textsc{ReFF} can simultaneously improve both format faithfulness (e.g., from 21.6\% in original LLaMA3 to 75.5\%) and general quality (e.g., from 47.3 to 61.6 F1 scores).

We further combine analyses and examples to explain how \textsc{ReFF} is able to obtain highly favorable results in terms of both format faithfulness and general quality. The discussion reveals that although often being consistent and aligned, format faithfulness and general quality of LLMs may also trade off as inversely correlated metrics. As a result, solely improving format faithfulness may cause LLMs generating well-formatted but semantically irrelevant content, while \textsc{ReFF} can combine the best to two worlds by involving both metrics.

Our main contributions are summarized as follows.

\begin{itemize}
    \item For a comprehensive evaluation of format faithfulness, we develop \textsc{FormatBench}, which covers a variety of tasks. Experiments show \textsc{FormatBench} is challenging for state-of-the-art LLMs.
    \item We propose \textsc{ReFF} by incorporating format checking in a reinforcement learning process. \textsc{ReFF} is validated to be highly effective in improving format faithfulness with or without extra training data.
    \item We offer an interpretability analysis to explain how \textsc{ReFF} can simutaneously improve both format faithfulness and general quality.
\end{itemize}

\section{Related Work}

\paragraph{Format-Related LLM Benchmarks}
In recent years, there has been significant attention paid to benchmarks and evaluation metrics in language modeling fields. Several notable benchmarks have been developed to evaluate the holistic effectiveness of LLMs \cite{wang2018glue, wang2019superglue, liang2022holistic, srivastava2023beyond}. Additionally, a few benchmarks have been proposed to evaluate format-related aspects. However, previous format-related benchmarks are task-specific, failing to provide a comprehensive evaluation of overall format faithfulness. For example, \textsc{Chem-*} \cite{skreta2023errors} exclusively addresses a domain-specific programming language, \textsc{Struc-Bench} \cite{tang2023struc} exclusively addresses text-to-table conversion, \textsc{IFEval} \cite{li2024instruction} exclusively addresses single-turn instruction, and \textsc{FoFo} \cite{xia2024fofo} exclusively addresses specific domain document generation. \textsc{FormatBench} differs from these benchmarks as it covers a variety of tasks, as is shown in Table \ref{tab:bench-compare}.

\begin{table*}[htbp]
    \centering
    \begin{tabular}{lll} 
    \toprule
    \textbf{Task} & \textbf{Format Requirements} & \textbf{Bad Cases} \\
    \hline
    \multirow{1}{*}{NER} & legal flat NER schema & \texttt{<PER>}Sarah{\textbf{\texttt{</ORG>}}} baked delicious cookies yesterday.\\
    \hline
    \multirow{2}{*}{CapSeg} & $\leq$ 42 characters per line & The shimmering lake reflected the colors of {\textbf{the setting sun.}} \texttt{<eob>} \\
    & $\leq$ 2 lines per block & The shimmering lake \texttt{<eol>} reflected the colors {\textbf{\texttt{<eol>}}} of the setting sun. \texttt{<eob>} \\
    \hline
    \multirow{1}{*}{MTT} & adherence to translation rules & \makecell[l]{{src:} Das Exanthem des M. Still ist ein Symptom von hoher Sensitivität. \\ {rule:} "Exanthem" should be translated into "rash" \\ The {\textbf{exanthema}} of Still's disease is a symptom of high sensitivity.} \\
    \hline
    XDL & successful compilation & \texttt{<!-- a piece of XDL code that} {\textbf{doesn't pass compilation}} \texttt{-->} \\
    \bottomrule
    \end{tabular}
    \caption{Core format requirements for four tasks in \textsc{FormatBench}, and examples of corresponding wrong responses. Parts of the response that do not meet the format requirements are shown by bolding.}
    \label{tab:bench-example}
\end{table*}

\paragraph{Format-Related LLM Adaptations}
Before the era of LLMs, controllable text generation (CTG) has been proposed to steer a model to generate desired texts according to given control conditions \cite{prabhumoye2020exploring, zhang2023survey}. However, CTG methods usually adopt specially designed finetuning schema or modify the sampling procedure in decoding steps \cite{miao2019cgmh, qin2022cold, kumar2022gradient}, which are intricate for LLMs. Recently, several works adopt prompt engineering \cite{skreta2023errors} or finetuning \cite{tang2023struc} to improve format following ability, but neglect the decidable nature of format problems. Unlike previous work, our proposed \textsc{ReFF} is designed based on the decidability of formats, and significantly outperforms previous approaches with or without extra training data.

\section{\textsc{FormatBench}}
\label{sec:tasks}

\textsc{FormatBench} is a collection of tasks with formatting requirements, as shown in the example in Figure \ref{fig:reff}. In this section, we will firstly introduce the variety that \textsc{FormatBench} covers, then outline the benchmark construction, and finally define the metrics associated with the benchmark.

\subsection{Variety}
\label{subsec:variety}

\textsc{FormatBench} endeavors to conduct a comprehensive evaluation of format faithfulness. To this end, it covers a variety of application scenes, human-computer interaction styles, and format requirement types.

\paragraph{Application Scene}
(1) The rigorous output format is necessary for various traditional NLP tasks. (2) In currently prevalent creative tasks, users often devise a format and ask LLMs to follow. (3) Promising LLM-based autonomous agents need to adhere a pre-defined format to interact with the environment. \textsc{FormatBench} combines the all three scenes.

\paragraph{Interaction Style} Apart from traditional format specifications in (1) single-turn instructions, \textsc{FormatBench} also involves evaluating format faithfulness in (2) multi-turn interactions, where an LLM iteratively receiving observations and choosing actions to meet the changing format requirements.

\paragraph{Format Types} Inspired by previous works \cite{li2024instruction, xia2024fofo}, \textsc{FormatBench} focuses on four aspects of format specifications, namely keyword inclusion, tag wrapping, length constraints, and coding. (1) Inclusion involves certain words to be included or excluded. (2) Wrapping involves enclosing a span of text with pre-defined tags or characters. (3) Length constrains the count of generated content, such as characters or sentences. (4) Coding here refers to more complex format structures that requires a compilers that synthesize the full text.

As is shown in Table \ref{tab:bench-compare}, \textsc{FormatBench} covers a variety of format-related tasks, and culminating a larger amount of test data compared to previous benchmarks.

\subsection{Construction}

\begin{figure}[htbp]
    \centering
    \includegraphics[width=0.9\linewidth]{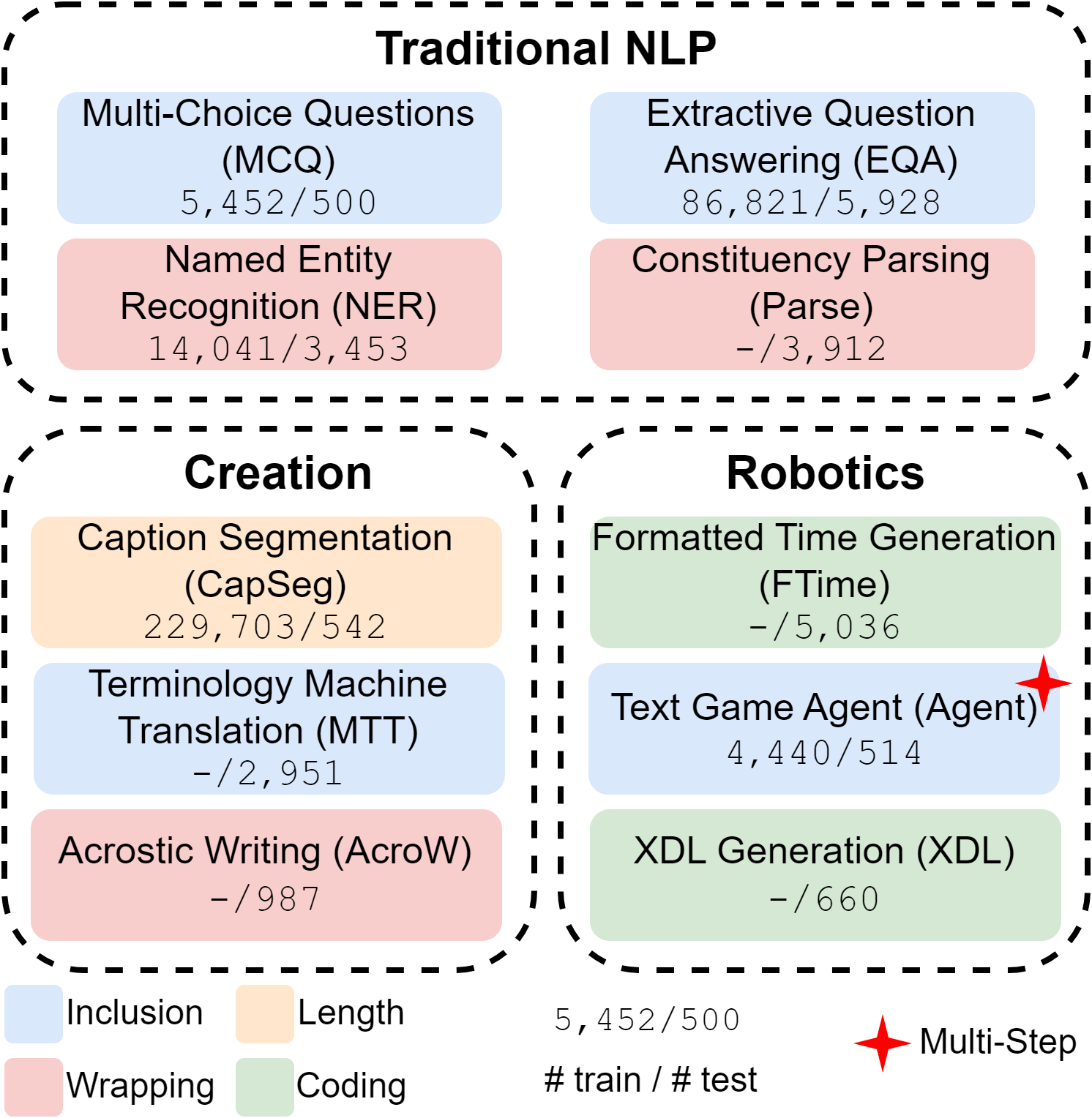}
    \caption{Tasks included in \textsc{FormatBench} with their corresponding groups and data sizes.}
    \label{fig:benchmark}
\end{figure}

\paragraph{Source}
Aiming at cover a variety of format-related tasks, the data construction in \textsc{FormatBench} (Figure \ref{fig:benchmark}) involves adaptation of existing datasets, collection of web data, and manual annotation. The task descriptions, data annotation, and quality control are detailed in Appendix \ref{app:benchmark}.

\paragraph{Format}
For each task, we define the format requirements based on previous literature and rough consensus. Some examples and their corresponding cases that fail to satisfy them are listed in Table \ref{tab:bench-example}, and all specific format requirements are listed in Appendix \ref{app:benchmark}. Moreover, we construct a corresponding format checker for each task in \textsc{FormatBench}. A format checker is a program, that given an input query and a response, determines whether the response adheres to the format requirements of the task. Formally, given a query $q$ and a generated response $r$, the format checker is defined as:

\begin{equation}
\label{equ:format-checker}
    \mathcal{F}(q, r) =
    \begin{cases}
        1 & \quad \text{if} \ r \ \text{fits the format},\\
        -1 & \quad \text{otherwise}.\\
    \end{cases}
\end{equation}

\subsection{Metrics}

We associate two metrics to \textsc{FormatBench}, namely, format faithfulness rate and general quality. Format faithfulness rate evaluates to what extent can an LLM $\mathcal{M}$ follows the format specifications, by calculating the format checker pass-rate across the set all samples $D$:

\begin{equation}
    FFR = 
    \mathbb{E}_{q \in D}[\mathds{1}(\mathcal{F}(q, \mathcal{M}(q)) = 1)].
\end{equation}

General quality of the generated responses is also evaluated, as a response is considered effective only if it is both faithful in format and correct in content. Due to the heterogeneity among all tasks in \textsc{FormatBench}, we respectively define the general quality metrics (e.g., BLEU, F1, accuracy) for each task, as described in Appendix \ref{app:benchmark}.

\section{Reinforcing Format Faithfulness}
\label{sec:methods}

\subsection{Algorithm}

Format problems have a decidable nature, where a format checker can easily discriminate whether generated texts adhere to format requirements or not. However, previous methods fail to fully take advantage of this feature.

We find that the format dicidability perfectly fits the reinforcement learning paradigm, where an environment provides rewards for the action given by an agent. To be specific, in a RL-based format faithfulness adaptation, LLMs can be viewed as agents that generating structured texts as actions, which are rewarded by an format checker environment.

In doing so, we proposed \textsc{ReFF} to use reinforcement learning for format faithfulness adaptation by rewarding models for correct formats and penalizing incorrect ones.

\begin{algorithm}[htbp]
\caption{\textsc{ReFF}}
\label{alg:reff}
\textbf{Input}: query set $Q$, format checker $\mathcal{F}$, LLM $\mathcal{M}$, \# epoch $n$\\
\textbf{Output}: adapted LLM $\mathcal{M}'$
\begin{algorithmic}[1] 
\STATE Let $\mathcal{M}' \gets \mathcal{M}$
\FOR{$epoch$ in $[1, 2, ..., n]$}
\FOR{$q$ in $Q$}
\STATE $r \gets \mathcal{M}'(q)$ \algorithmiccomment{response generation}
\STATE $s \gets \mathcal{F}(q, r)$ \algorithmiccomment{format checking, $s \in \{-1, 1\}$}
\STATE $\mathcal{M}' \gets step(\mathcal{M}', q, r, s)$ \algorithmiccomment{PPO stepping}
\ENDFOR
\ENDFOR
\STATE \textbf{return} $\mathcal{M}'$
\end{algorithmic}
\end{algorithm}

The process of \textsc{ReFF} is shown in Algorithm \ref{alg:reff}, where the $step()$ is the function of reinforcement learning from human feedback style (RLHF-style) stepping aimed at updating the LLM given the action and the reward. The used RLHF-style loss function \cite{ziegler2019fine, ouyang2022training} differs from the original proximal policy optimization (PPO) \cite{schulman2017proximal}, in that it additionally adds a Kullback-Leibler (KL) penalty from the original model to prevent the adapted model from shifting too far. The algorithm generally shares the same procedure with RLHF, except that the computation of rewards is different. Specifically, RLHF often uses a pre-trained reward model, while \textsc{ReFF} relies on format checkers to compute the rewards. We offer a rigorous math representation of \textsc{ReFF} algorithm in Appendix \ref{app:algorithm}.

\subsection{Settings}

Considering the data availability in various real-world scenarios, we set three settings for RL in \textsc{ReFF}, whose accessible data groups (query set $Q$ in Algorithm \ref{alg:reff}) are list in Table \ref{tab:reff-settings}.

\begin{table}[htbp]
    \centering
    \begin{tabular}{lccc} 
    \toprule
    \textbf{Settings} & \makecell[c]{\textbf{Test} \\ \textbf{Queries}} & \makecell[c]{\textbf{Train} \\ \textbf{Queries}} & \makecell[c]{\textbf{Train} \\ \textbf{Labels}} \\
    \hline
    \textsc{ReFF}-tst & \CheckmarkBold & \XSolidBrush & \XSolidBrush \\
    \textsc{ReFF}-trn & \XSolidBrush & \CheckmarkBold & \XSolidBrush \\
    \textsc{ReFF}-trn-ft & \XSolidBrush & \CheckmarkBold & \CheckmarkBold \\
    \bottomrule
    \end{tabular}
    \caption{Data used for RL in three settings of \textsc{ReFF}.}
    \label{tab:reff-settings}
\end{table}

\paragraph{Test-Only \textsc{ReFF}}
When there exists no extra training data, LLMs can use queries in the test set as the query set $Q$. Notably, no label of the test set is available to the model in this setting. However, this setting only applies to the offline scenarios, where LLMs handle a batch of queries and generate all responses subsequently.

\paragraph{Train-Only \textsc{ReFF} w./wo. Finetuning}
Train-only setting can be applied in an online scenario, where the queries are processed and responsed one by one, as the adaptation of LLMs only involves training queries as the query set $Q$. Additionally, considering that a training set often includes both queries and labels, we further study a train-only with finetuning setting, where the reinforcement process is implemented after finetuning on the training set.

\section{Experiments}
\label{sec:exps}

\begin{table*}[htb!]
    \centering
    \begin{tabular}{lccccccccccc}
        \toprule
        \multicolumn{1}{c}{\textbf{Models}} & \rotatebox{45}{\textbf{MCQ}} & \rotatebox{45}{\textbf{EQA}} & \rotatebox{45}{\textbf{NER}} & \rotatebox{45}{\textbf{Parse}} & \rotatebox{45}{\textbf{CapSeg}} & \rotatebox{45}{\textbf{MTT}} & \rotatebox{45}{\textbf{AcroW}} & \rotatebox{45}{\textbf{FTime}} & \rotatebox{45}{\textbf{Agent}}  & \rotatebox{45}{\textbf{XDL}} & \textbf{avg.}\\
        \hline
        GPT-3.5      & 99.0 & 89.7 & 95.3 & 36.2 & 45.8 & 56.0 & 44.5 & 95.4 & 71.0 & 5.5 & 63.8 \\
        LLaMA3       & 97.0 & 89.6 & 84.8 & 0.2  & 21.6 & 52.3 & 1.7 & 99.4 & 88.3 & 13.3 & 54.8 \\
        Gemma        & 98.0 & 90.0 & 82.0 & 5.5  & 28.2 & 50.9 & 2.0 & 98.8 & 91.4 & 0.0  & 54.7 \\
        Qwen1.5      & 96.6 & 91.3 & 71.9 & 4.6  & 24.7 & 55.3 & 0.9 & 99.0 & 88.1 & 10.3 & 54.3 \\
        Mistral      & 96.0 & 91.1 & 86.2 & 1.6  & 34.1 & 40.4 & 6.9 & 99.4 & 86.8 & 0.0 & 54.2 \\
        Mistral-inst & 96.0 & 89.5 & 77.5 & 1.4  & 32.1 & 54.2 & 3.0 & 97.2 & 79.0 & 0.0  & 53.0 \\
        LLaMA2       & 97.4 & 86.9 & 83.9 & 0.3  & 25.5 & 39.9 & 0.1 & 99.8 & 73.7 & 8.9 & 51.6 \\
        LLaMA        & 89.6 & 88.5 & 74.1 & 0.3  & 22.1 & 29.7 & 0.0 & 72.8 & 81.7 & 42.4 & 50.1 \\
        Falcon       & 83.4 & 77.8 & 62.7 & 0.1  & 26.9 & 20.5 & 0.0 & 32.0 & 63.2 & 45.5 & 41.2 \\
        Falcon-inst  & 83.2 & 55.4 & 26.0 & 0.0  & 22.7 & 11.9 & 0.1 & 35.9 & 35.4 & 54.5 & 32.5 \\
        \bottomrule
    \end{tabular}
    \caption{Format faithfulness rate (\%) of original models on \textsc{FormatBench}.}
    \label{tab:ff-scores}
\end{table*}

\subsection{Baselines}

There are two groups of baselines we implement to compare with \textsc{ReFF}, namely, refinement and finetuning.

\paragraph{Refinement}
There exists many works on prompt engineering about augmenting LLMs with internal reflections \cite{wei2022chain, madaan2023self} to refine their initial content. Among them, a recent paper focuses on generating well-structured codes \cite{skreta2023errors}. Inspired by this, we take refinement as a general prompt schema for improving format faithfulness on all tasks. Generally, an LLM iteratively polishes the output format according to error information from the format checker. Optionally, we further augment the refinement process with LLM internal thoughts following the CoT \cite{wei2022chain} and ReAct \cite{yao2023react} prompting.

\paragraph{Finetuning}
The abilities of LLMs can be further adapted according to specific goals by finetuning \cite{zhao2023survey}. Specifically, recent work \cite{tang2023struc} finetunes LLMs to generate well-structured data on specific tasks. Inspired by these works, we propose to conduct finetuning on LLMs to improve the overall format faithfulness.

\subsection{Experimental Setup}

\paragraph{Models}
We conduct a comprehensive evaluation on format faithfulness with \textsc{FormatBench} across many state-of-the-art open-source LLMs sizing about 7B, including LLaMA-7B, LLaMA-2-7B, LLaMA-3-8B, Qwen-1.5-7B, Falcon-7B, Falcon-7B-Inst, Mistral-7B-v0.3, Mistral-7B-Inst-v0.3, Gemma-7B. We further compare these models to closed-source GPT-3.5 (gpt-3.5-turbo-instruct). Note that we only use instruction models, as all tasks in \textsc{FormatBench} are instruction tasks, where instruction prompting styles are more suitable and flexible than chat ones. In adaptation experiments including refinement, finetuning, and \textsc{ReFF}, we use LLaMA-3-8B as the base model, as it exhibits a favorable format faithfulness in the original model evaluation.

\paragraph{Adaptation Implementation}
We use \texttt{trl} \cite{vonwerra2022trl} library to implement the finetuning and the RLHF-style PPO of \textsc{ReFF}. More information about the implementation of adaptation methods are detail in Appendix \ref{app:implementation}.

\paragraph{Hyper-Parameters}
To ensure the robustness and reliability of the results, we try to use default and commonly-used hyper-parameters, and keep them consistent among different experiments. Here we list several key points, and the detailed hyper-parameters are outlined in Appendix \ref{app:implementation}.
\begin{itemize}
    \item In generation, we adopt greedy decoding in all experiments for a fair and efficient comparison.
    \item We use LoRA \cite{hu2021lora} in all LLM adaptation experiments with a consistent configuration $r=16$.
    \item In fintuning, we use a constant learning rate $2e-5$ and train for $3$ epochs with $256$ instances per batch.
    \item In reinforcement learning, we set target of KL divergency to be $6$, use a constant learning rate $1.41e-5$, and train for $3$ epochs with $32$ instances per batch.
\end{itemize}

\begin{table*}[htb!]
    \centering
    \begin{tabular}{lccccccc}
        \toprule
        \multirow{2}{*}{\textbf{Models}} & \multicolumn{4}{c}{\textbf{Format Faifulness Rate} ($\uparrow$)} & \multicolumn{3}{c}{\textbf{General Quality} ($\uparrow$)} \\
        \cmidrule(lr){2 - 5} \cmidrule(lr){6 - 8} & {\textbf{NER}} & {\textbf{CapSeg}} & {\textbf{MTT}} & {\textbf{XDL}} & {\textbf{NER}} & {\textbf{CapSeg}} & {\textbf{MTT}}\\
        \hline
        GPT-3.5        & 95.3 & 45.8 & 56.0 & 5.5  & 94.3 & 40.6 & 30.9 \\
        \ \ + refine   & 96.1 & 62.5 & 73.1 & 5.9  & 94.3 & 42.0 & 31.7 \\
        \ \ + refine*  & 96.7 & 72.3 & 83.8 & 5.9 & 94.2 & 23.0 & 31.5 \\
        \hline
        LLaMA3         & 84.8 & 21.6 & 52.3 & 13.3 & 88.3 & 47.3 & 32.2 \\
        \ \ + refine   & 85.0 & 21.8 & 61.8 & 13.3 & 88.3 & 47.2 & 17.7 \\
        \ \ + refine*  & 85.6 & 33.0 & 69.2 & 13.3 & 88.5 & 21.8 & 13.7 \\
        \hline
        ReFF-tst-NER & 96.7 & 20.7  & 58.5 & 17.6 & 91.4 & \textbf{47.8} & 33.0 \\
        ReFF-tst-CapSeg  & 87.6 & 95.0  & 52.0 & 15.0 & 87.9 & 46.4 & 31.3 \\
        ReFF-tst-MTT     & 88.8 & 21.6  & \textbf{98.2} & 13.0 & 88.4 & 47.6 & 31.0 \\
        ReFF-tst-XDL  & 86.2 & 21.4  & 51.0 & \textbf{52.6} & 89.0 & 47.3 & 31.9 \\
        ReFF-tst     & \textbf{99.7} & \textbf{100.0} & 97.2 & 14.8 & \textbf{93.1} & 40.9 & \textbf{35.3} \\
        \bottomrule
    \end{tabular}
    \caption{Format faithfulness rate (\%) and general quality (F1 for NER and CapSeg, BLEU-4 for MTT) in the test-only setting. Best results among LLaMA3-based models are bolded. \textsc{ReFF}-tst-\texttt{[task]} refers to the model adapting with exclusively corresponding dataset, while \textsc{ReFF}-tst mixes and shuffles all four datasets. The asterisk symbol denotes refinement with internal thoughts. The general quality of XDL is not evaluated due to the need for a high level of expert knowledge, as detailed in Appendix \ref{app:benchmark}.}
    \label{tab:offline}
\end{table*}

\subsection{Original Model Results}

We evaluate the models using the prompts in Appendix \ref{app:prompts}.

\paragraph{Results}
The format faithfulness results of original LLMs on \textsc{FormatBench} are presented in Table \ref{tab:ff-scores} (general quality in Appendix \ref{app:gq}). We find the benchmark to be both discriminating and challenging for LLMs. Firstly, it can be observed that stronger model like GPT-3.5 does exhibits better faithfulness to format, as its format faithfulness rates surpasses those of other smaller open-source models. Secondly, it is validated that format tasks are still highly challenging for even the most capable models.

\paragraph{Outliers}
Moreover, there are some intriguing exceptions found in the results, where smaller models like Falcon-inst demonstrate superior faithfulness compared to GPT-3.5 in XDL task (54.5\% versus 5.5\%), as shown in Table \ref{tab:ff-scores}. We try to explain this phenomenon in Section \ref{sec:analysis} by studying the relation between format faithfulness and general quality.

\subsection{Adapted Model Results}

In order to adapt LLMs to alleviate format unfaithfulness, we propose \textsc{ReFF} by incorporating a format checker into a reinforcement learning process. We further offer three settings in Table \ref{tab:reff-settings}, namely \textsc{ReFF}-tst, \textsc{ReFF}-trn, and \textsc{ReFF}-trn-ft to cover different application scenarios.

We study the adaptation approaches in test-only setting with four tasks including NER, CapSeg, MTT, and XDL. The corresponding examples are listed in Table \ref{tab:bench-example}. These four tasks are chosen for two reasons, (1) they fully covers the application scenes and format types in our proposed taxonomy shown in Figure \ref{fig:benchmark}, and (2) their format requirement difficulties are moderate as shown in Table \ref{tab:ff-scores}. In the train-only setting, we choose the NER and CapSeg tasks as the other two tasks are not attached with training data.

\paragraph{\textsc{ReFF}-tst}
The results of \textsc{ReFF}-tst and other baselines in test-only setting are shown in Table \ref{tab:offline}. Comparing \textsc{ReFF} to other approaches, it is obvious that \textsc{ReFF} can significantly improve the format faithfulness rate while keep the general quality comparable without any annotated data. Notably, the format faithfulness of \textsc{ReFF} exceeds not only its LLaMA3 baselines, but also GPT-3.5, validating the high effectiveness of our proposed \textsc{ReFF} approach. Moreover, comparing \textsc{ReFF} trained on one specific task (\textsc{ReFF}-tst-\texttt{[task]}) to that trained on mixed data (\textsc{ReFF}-tst), we can find that the catastrophic forgetting phenomenon is not significant in format-related reinforcement learning, as the format faitfhfulness rate of one task doesn't significantly drops when combined with data from other tasks.

\paragraph{\textsc{ReFF}-trn}
When test data is not available beforehand, while there exists training data for adaptation, \textsc{ReFF}-trn succeeds in improving format faithfulness to the extent that \textsc{ReFF}-tst does, as is shown in Table \ref{tab:online}. These results proves the format faithfulness improvement is robust, and does not result from overfitting to the test set.

\paragraph{\textsc{ReFF}-trn-ft}
By combining reinforcement for improving format faithfulness and finetuning for improving general quality, \textsc{ReFF}-trn-ft obtains highly favorable results on both metrics, as is shown in Table \ref{tab:online}.

\begin{table}[htb!]
    \centering
    \begin{tabular}{lcccc}
        \toprule
        \multirow{2}{*}{\textbf{Models}} & \multicolumn{2}{c}{\textbf{FF Rate} ($\uparrow$)} & \multicolumn{2}{c}{\textbf{GQ} ($\uparrow$)} \\
        \cmidrule(lr){2 - 3} \cmidrule(lr){4 - 5} & {\textbf{NER}} & {\textbf{CapSeg}} & {\textbf{NER}} & {\textbf{CapSeg}}\\
        \hline
        GPT-3.5          & 95.3 & 45.8 & 94.3 & 40.6 \\
        \ \ + refine     & 96.1 & 62.5 & 94.3 & 42.0 \\
        \ \ + refine*    & 96.7 & 72.3 & 94.2 & 23.0 \\
        \hline
        LLaMA3           & 84.8 & 21.6 & 88.3 & 47.3 \\
        \ \ + finetune   & 99.0 & 38.2 & \textbf{95.9} & \textbf{63.6} \\
        \hline
        ReFF-trn         & \textbf{99.8} & \textbf{99.8} & 92.6 & 40.9 \\
        ReFF-trn-ft      & 99.2 & 75.5 & 95.2 & 61.6 \\
        \bottomrule
    \end{tabular}
    \caption{Format faithfulness rate (\%) and general quality (F1) in the train-only setting. Best results are bolded. The asterisk symbol denotes refinement with internal thoughts.}
    \label{tab:online}
\end{table}

\paragraph{Outliers}
Moreover, similar to the exceptions in the original model results, there also exists outliers in adapted model results. As shown in Table \ref{tab:offline}, the general quality drops drastically with refinement on LLaMA3 MTT task (from 32.2 to 13.7 in BLEU-4), and \textsc{ReFF}-tst-XDL obtains an unusually higher format faithfulness rate (52.6\%) on XDL task than its counterpart trained with all four tasks (14.8\%). We will also give an explanation to the outliers in Section \ref{sec:analysis}.

\begin{figure}[tbp]
    \centering
    \includegraphics[width=\linewidth]{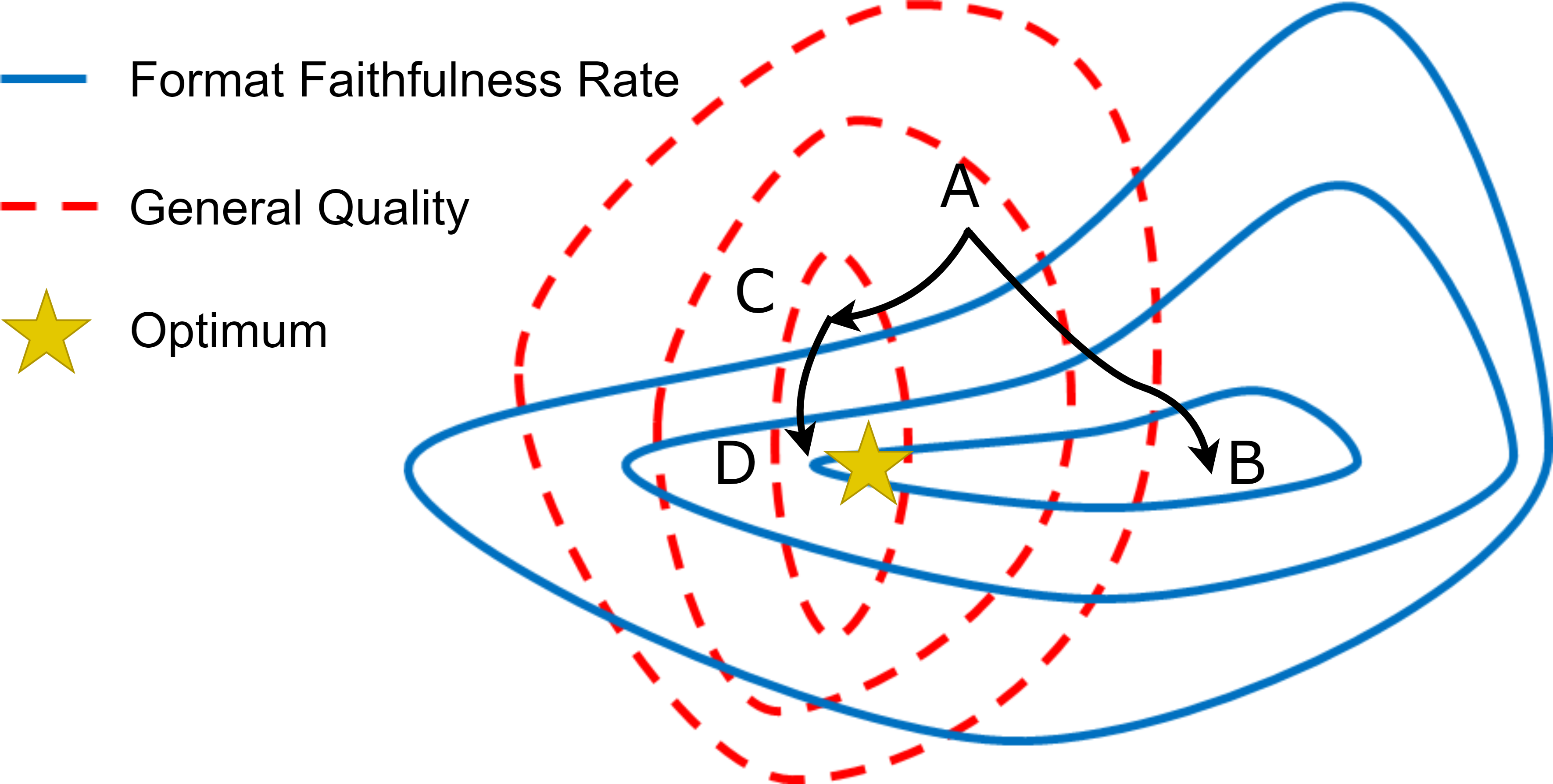}
    \caption{Conceptual contour map of format faithfulness and general quality. Inner circles indicate higher scores for both metrics. Solely improving format faithfulness (A $\rightarrow$ B) may result in an LLM with high format faithfulness but low general quality. \textsc{ReFF} can get the best of two worlds by combining finetuning (A $\rightarrow$ C) and reinforcement (C $\rightarrow$ D).}
    \label{fig:contour}
\end{figure}

\section{Analysis}
\label{sec:analysis}

\begin{figure*}[htb!]
    \centering
    \includegraphics[width=\linewidth]{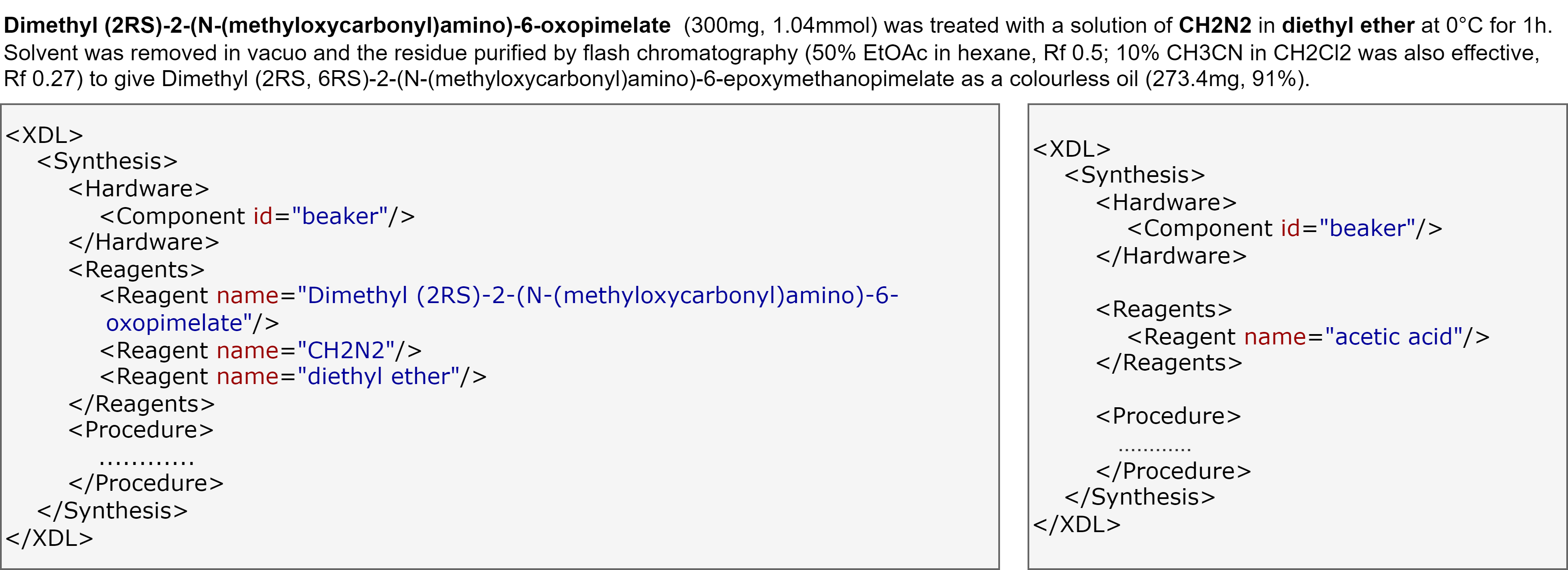}
    \caption{An instance in XDL task (top), the corresponding response of \textsc{ReFF}-tst (left), and that of \textsc{ReFF}-tst-XDL which obtains a higher format faithfulness rate on XDL task (right). \textsc{ReFF}-tst-XDL generates syntactically correct but irrelevant code.}
    \label{fig:xdl-examples}
\end{figure*}

We begin our analysis by considering the outliers in the results of experiments in Section \ref{sec:exps}, including:

\begin{itemize}
    \item Outlier 1: Although considered being more capable, GPT-3.5 does not perform as well as 7B open-source models in format faithfulness on XDL task (Table \ref{tab:ff-scores}). 
    \item Outlier 2: When using refinement techniquesin LLaMA3, format faithfulness improves significantly, but general quality suffers from a drastic drop on MTT task (Table \ref{tab:offline}).
\item Outlier 3: \textsc{ReFF}-tst-XDL significantly outperforms \textsc{ReFF}-tst, its counterpart with similar training data, in format faithfulness rate on XDL task (Table \ref{tab:offline}).
\end{itemize}

These exceptions have a same pattern, that is, the discrepancy between format faithfulness and general quality. More specifically, a model that is supposed to have higher quality may exhibits poorer format faithfulness. By analyzing the three outliers and the underlying shared pattern, we summarize the relation between format faithfulness and general quality, and then explain how our proposed \textsc{ReFF} shows highly favorable results in both metrics.

In this section, we will first offer an insightful illustration to explain the discrepancy between format faithfulness and general quality, and then discuss a specific case of Outlier 3.

\subsection{Illustration}

Figure \ref{fig:contour} illustrates the conceptual relation between format faithfulness and general quality. Although being consistent in most scenarios, format faithfulness and general quality are two different metrics, and sometimes may trade off as inversely correlated indicators.

An LLM can obtain an acceptable general quality, while failing to adhere format requirements, as point A shows. Meanwhile, an LLM can be completely faithful to format, while being poor in general quality, as point B shows. Solely adapting LLMs with the supervision of format faithfulness rates may guide them from the former situation (point A) to the latter one (point B).

The discussion above give all three exceptions an explanation. Point A explains Outlier 1, where strong LLMs like GPT-3.5 generate poor-structured texts. Point B explains Outlier 3, where an LLM may gain a significant format faithfulness improvement without higher general quality. The trace from Point A to Point B explains Outlier 2, where an LLM compromises general quality to improve format faithfulness.

Fortunately, experiments in Table \ref{tab:online} show that our proposed \textsc{ReFF} is able to combine the best of both worlds. In the finetuning process in \textsc{ReFF}, an LLM is firstly adapted to a position with high general quality (point A to C). In the following, it is futher adapted for better format faithfulness by reinforcement  (point C to D). With the application of KL regularization term that prevents the model from shifting too far from the original parameters (point C) in reinforcement process, the LLM after reinforcement (point D) can avoid significant decrease in general quality.

\subsection{Examples of XDL Task}

To explain why in Outlier 3 the XDL task falls into the point B (high format faithfulness, low general quality), we take a closer look at the example in Figure \ref{fig:xdl-examples}. In the example, \textsc{ReFF}-tst-XDL sneakily passes the format checker by generating short and simple well-formatted code (high format faithfulness) that is irrelevant to the instruction (low general quality). The phenomenon is an typical instance of mode collapse in RLHF, characterized by a reduced diversity in produced samples \cite{casper2023open}. RL-based adaptation techniques that may suffer from the mode collapse still have some room for substantially improving the performance in XDL task, as they require the original LLM to produce a number of diverse correctly formatted responses for rewarding.

\section{Conclusion}

In this paper, we aim to conduct a comprehensive evaluation of format faithfulness and enhance it without compromising the general quality of LLMs. In doing so, we firstly propose \textsc{FormatBench}, a format-related benchmark that covers a variety of tasks. \textsc{FormatBench} is shown to be highly discriminating and challenging for state-of-the-arts LLMs. Subsequently, by utilizing the decidable nature of formats, we incorporate format checking procedures into reinforcement learning to propose \textsc{ReFF}. Extensive experiments validate the high effectiveness of \textsc{ReFF} in simultaneously enhancing both format faithfulness and general quality. Finally, we provide an interpretability analysis to elucidate the reasons behind \textsc{ReFF}'s effectiveness by exploring the relationship between format faithfulness and general quality.

\section*{Acknowledgements}

We’d like to thank all the anonymous reviewers for their diligent efforts in helping us improve this work. This work is supported by the National Natural Science Foundation of China (Grant No. U21B2009) and Beijing Institute of Technology Science and Technology Innovation Plan (Grant No. 23CX13027).

\bibliography{custom}

\clearpage
\appendix

\section{Details of \textsc{FormatBench}}
\label{app:benchmark}

\subsection{Task Descriptions}

\paragraph{MCQ}
The Text TRtrieval Conference (TREC) question classification \cite{li-roth-2002-learning, hovy-etal-2001-toward} is a task that, given a question, maps it to one of the given classes, which provides a semantic constraint on the sought-after answer.

\paragraph{EQA}
Stanford Question Answering Dataset (SQuAD) \cite{rajpurkar2016squad} is a reading comprehension dataset, where the answer to every question is a segment of text, or span, from the corresponding reading passage. We use a copied format in this task, i.e., requiring LLMs to directly copy the span of the passage without modification.

\paragraph{NER}
CoNLL-2003 \cite{sang2003introduction} is a named entity recognition (NER) task to detect and categorize named entities.

\paragraph{Parse}
We conduct constituency parsing on the open source subset of the Penn Treebank (PTB) \cite{marcus1993building} using the bracket sequence representation of a constituency tree.

\paragraph{CapSeg}
We adapt MuST-Cinema \cite{karakanta-etal-2020-must} dataset, a multilingual speech translation corpus built from TED subtitles, to construct the caption segmentation. CapSeg involves inserting end-of-block and end-of-line tags in the raw English text to represent the split of captions in videos, thus simulating the generation of English video captions.

\paragraph{MTT}
WMT 2023 Terminology Shared Task \cite{semenov2023findings} is a Germany-English machine terminology translation (MTT) task that challenges machine translation systems to accurately and effectively translate technical terms and specialized vocabulary.

\paragraph{AcroW}
An acrostic poem is a literary form in which the initial letter of each line is arranged to spell out a hidden message. We combine existing datasets and acrostic poems crawled from the Internet. In this task, an LLM is challenged to compose an acrostic poem, adhering to the format of having the first letter of each line spell out the intended message.

\paragraph{FTime}
Formatted Time Generation (FTime) task is to generate formatted time representations based on natural language instructions. This task holds particular relevance in reminder applications, which involve the translation of natural languages into formatted time strings. Examples of FTime task are provided in Table \ref{tab:ftime}, where the illustrations of both single-trigger and repetitive-trigger time format are displayed.
\begin{table}[htb!]
    \centering
    \begin{tabular}{lp{6.1cm}}
    \hline
    \textbf{Ref.} & \scalebox{0.93}{\texttt{20021019T140000:Saturday}}\\
    \textbf{Inst.} & \scalebox{0.93}{... soup will be ready in 20 minutes ...}\\
    \textbf{Res.} & \scalebox{0.93}{\texttt{20021019T142000}}\\
    \hline
    \textbf{Ref.} & \scalebox{0.93}{\texttt{20121215T090000:Saturday}} \\
    \textbf{Inst.} & \scalebox{0.93}{... walk my dog at 10 a.m. every Monday.}\\
    \textbf{Res.} & \scalebox{0.93}{\texttt{R-1/20121217T100000/P0Y0M7DT0H0M0S}} \\
    \hline
    \end{tabular}
    \caption{In FTime, an LLM is to generate a time string (Res.) referred to by the instruction (Inst.), assuming that the instruction is issued at the reference time (Ref.). The output should be in accordance with the single-trigger (top) or repetitive-trigger (bottom) format.}
    \label{tab:ftime}
\end{table}

\paragraph{Agent}
The First TextWorld Problems (FTWP) \cite{ftwp} is to build an AI agent that moves automatically and efficiently in a text simulated world according to text feedback, and finally completes the given goal.

\paragraph{XDL}
XDL (Chemical Description Language) \cite{seifrid2022autonomous} is an XML-based programming language used in chemical synthesis specification and experimental procedure transfer among robots and laboratories. Following previous work \cite{skreta2023errors}, we conduct XDL Generation task by examining the ability of LLMs to generate compilable XDL programs given the description of XDL.

\subsection{Annotation and Quality Control}

\begin{figure*}[htb!]
    \centering
    \includegraphics[width=0.75\linewidth]{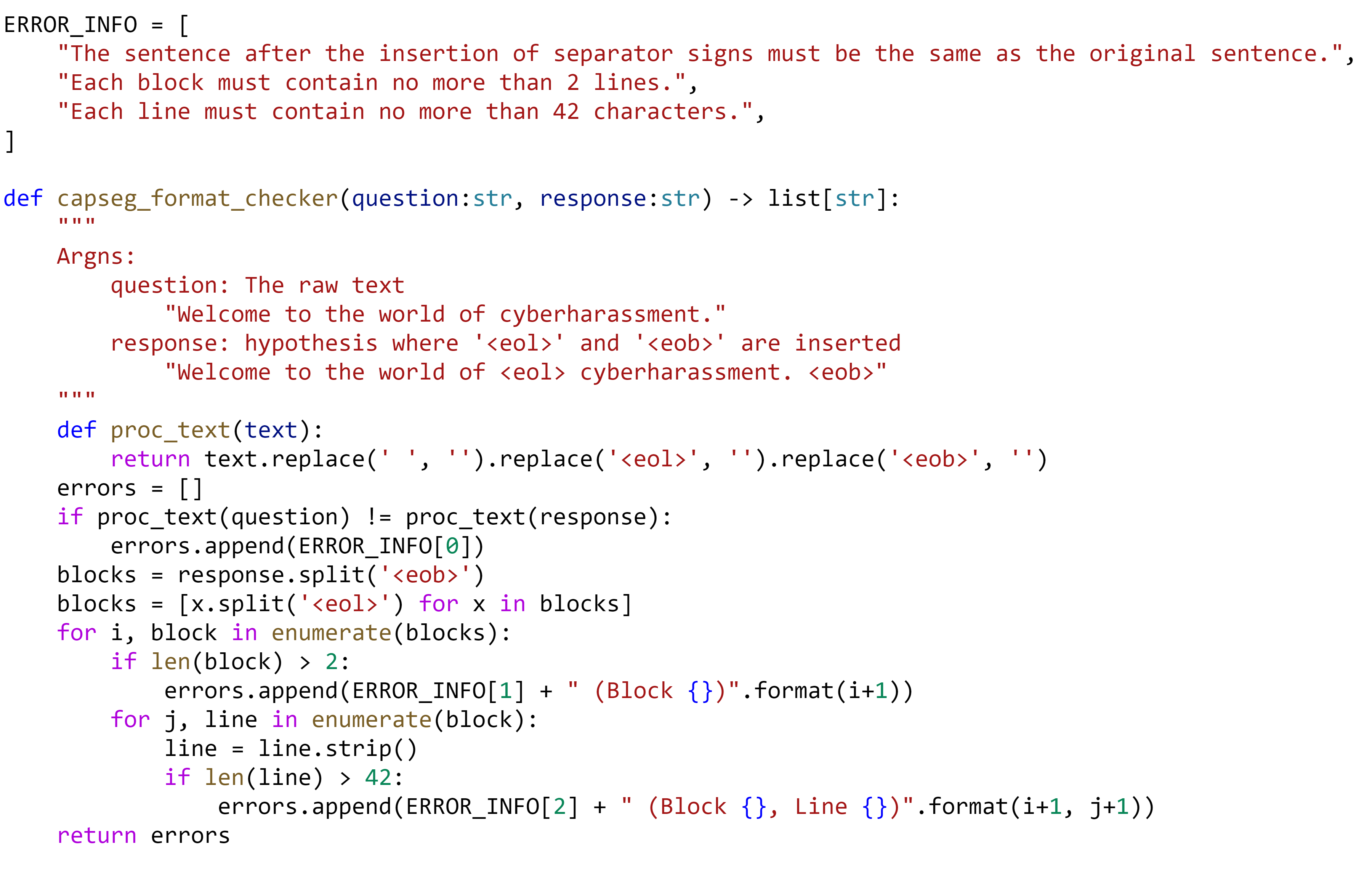}
    \caption{The format checker for CapSeg task.}
    \label{fig:format-checker}
\end{figure*}

\paragraph{AcroW}
We collect the acrostic poem dataset from Poem Hunter\footnote{\url{https://www.poemhunter.com/}}, focusing on the acrostic category to gather 927 acrostic poems via web scraping. We additionally combine Kaggle Poems Dataset\footnote{\url{https://www.kaggle.com/datasets/michaelarman/poemsdataset}} acrostic poems with these data. To ensure data quality and consistency, we eliminate redundant punctuation and standardize poem lines, retaining only the acrostic portion. Moreover, we filter out poems that do not meet acrostic requirements, such as initial letters failing to form coherent words or lacking relevance, to maintain data accuracy. The processed dataset undergoes manual inspection to verify quality and integrity, resulting in a dataset of 987 valid acrostic poems.

\paragraph{FTime}
Following ISO 8061 standard\footnote{\url{https://en.wikipedia.org/wiki/ISO_8601}}, we define the time format for FTime and categorize them into non-recurring time format and recurring time format. The non-recurring format is represented as "YYYYMMDDTHHMMSS", and the recurring format is represented as "Rn/YYYYMMDDTHHMMSS/PnYnMnDTnHnMnS".

In the definition of FTime task, we categorize the instructions into three classes. In the first category, the provided instruction contains an event interval (such as "remind me in 20 minutes"), and the final result is the reference time plus this time interval. In the second category, the instruction provided contains a specific time (such as "tomorrow morning at 8 o'clock"). The final result needs to be obtained based on the reference time and the time in the instruction. In the third category, the instruction provided includes a recurring event (such as "at 10 a.n. every Monday"). The final outcome requires determining the number of recurrences, the recurrence interval, and identifying the time of the first event trigger based on the reference time. The format in the first and second categories forms a non-recurring time, which that in the third category constitutes a recurring time. An LLM is to generate a time representation in accordance to the format based on the reference time, the given instruction, and the category to which the instruction pertains.

In the construction of FTime, three components need to be considered, namely reference time, natural language instruction, and the result. All the reference times and their corresponding category tags are automatically generated in advance. We compose the instruction part through two methods, including slot filling and manual composition. For slot filling method, we first design templates with placeholders for a time slot and an event slot. Then, for each of the three classifications mentioned above, we prepare a set of events to be filled in. Subsequently, several templates are randomly selected for each event, and the event slot is filled accordingly. In this way, we generated 4536 instruction instances. To diversify the data, we manually compose another 500 instruction instances. These manually composed data are not limited by the slot filling framework, thus obtaining better flexibility and higher difficulty. Combining the two parts together, we develop 5036 pieces of instruction in total. Given a reference time and a instruction, the corresponding result is annotated manually.

After annotation, we randomly sample 3\% data to conduct a cross-validation involving re-annotation by a different annotator. The consistency between the initial annotation and the re-annotation is found to be 98.01\%, thereby confirming the trustworthiness of our annotated data.

\subsection{Format Requirements and Checkers}

Based on previous literature and rough consensus, we design a group of format requirements for each task in \textsc{FormatBench}, as is shown in Table \ref{tab:format-requirements}.

\begin{table}[htb!]
    \centering
    \begin{tabularx}{\linewidth}{lX}
        \toprule
        \textbf{Tasks} & \textbf{Requirements} \\
        \hline

        \multirow{2}{*}{MCQ} & The answer should be among the legal class options. \\
        \hline

        \multirow{2}{*}{EQA} & The answer should be a segment of text, or span, from the corresponding reading passage.\\
        \hline

        \multirow{4}{*}{NER} & Every opening tag must be closed with a corresponding closing tag.\\
        \cmidrule(lr){2 - 2} & The sentence after the insertion of NER tags must be the same as the original sentence.\\
        \hline

        \multirow{10}{*}{Parse} & The final outcome should be a string with properly closed parentheses.\\
        \cmidrule(lr){2 - 2} & The sentence after the insertion of brackets and tags must be the same as the original sentence.\\
        \cmidrule(lr){2 - 2} & Use legal labels exclusively.\\
        \cmidrule(lr){2 - 2} & Use word-level labels exclusively for a word.\\
        \cmidrule(lr){2 - 2} & Use clause-level or phrase-level labels exclusively for a text span.\\
        \cmidrule(lr){2 - 2} & There must exist subtrees or words that are specific to a given label.\\
        \hline
        
        \multirow{5}{*}{CapSeg} & The sentence after the insertion of separator signs must be the same as the original sentence.\\
        \cmidrule(lr){2 - 2} & Each block must contains no more than 2 lines.\\
        \cmidrule(lr){2 - 2} & Each line must contains no more than 42 characters.\\
        \hline

        \multirow{2}{*}{MTT} & Each source term should be translated according to the terminology translation rules.\\
        \hline

        \multirow{2}{*}{AcorW} & The first letter of each line should spell the given string.\\
        \hline

        \multirow{5}{*}{FTime} & The result should comply with the YYYYMMDDTHHMMSS or Rn/YYYYMMDDTHHMMSS/PnYnMnDTnHnMnS format.\\
        \cmidrule(lr){2 - 2} & The date given in the result should form a legal date.\\
        \hline

        \multirow{2}{*}{Agent} & The result should be a legal action that can be executed in the simulated environment.\\
        \hline

        \multirow{2}{*}{XDL} & The result should be a piece of XDL code that successfully compiles.\\
        \bottomrule
    \end{tabularx}
    \caption{Format requirements for tasks in \textsc{FormatBench}.}
    \label{tab:format-requirements}
\end{table}

Besides, we develop one format checker for each task to determine whether a response violate any format requirement. A format checker, which is theoretically a recognizer of formal languages, is implemented as a Python program that check the format requirements that needs to be satisfied one by one.

Moreover, for the convenience of refinement format engineering, we additionally let the format checkers return the error messages, i.e., which part of the response violates which requirement. Notably, the Agent and the XDL task can rely on the simulated environment and the compiler to generate error messages respectively.

An example of format checker is shown in Figure \ref{fig:format-checker}.

\subsection{General Quality Metrics}

\paragraph{MCQ}
We calculate the accuracy, which is the percentage of correctly predicted instances out of the total instances in the test set, to evaluate classification performance in the task.

\paragraph{EQA}
We calculate the F1 score to measure the average overlap between the prediction and ground truth as previous works \cite{rajpurkar2016squad}.

\paragraph{NER}
Similar to QA tasks, we use the F1 score by treating the prediction and the ground truth as bags of words \cite{sang2003introduction}.

\paragraph{Parse}
We calculate the F1 score as previous study\footnote{\url{https://www.cs.princeton.edu/courses/archive/fall19/cos484/lectures/lec10.pdf}}. Combining symbol, beginning, and ending into a constituent, we can get candidate brackets and gold standard brackets in prediction and ground truth respectively.

\paragraph{CapSeg}
As previous works \cite{karakanta-etal-2020-must}, we calculate the F1 score by counting the correct\_breaks and the total\_breaks in prediction and ground truth respectively.

\paragraph{MTT}
We calculate the BLEU-4 score between the reference translation and the hypothesis for MTT task evaluation.

\paragraph{AcroW}
Following previous work \cite{agarwal2020acrostic}, we utilize GPT-4-0125-preview to score generated acrostic poems based on four aspects: poetic essence, rhyme, content, and readability. Each aspect is scored from 0 to 5, with 0 being the lowest and 5 being the highest. 
\begin{itemize}
    \item Poetic essence: Does the poem embody the spirit of poetry and feel like a genuine piece of verse?
    \item Rhyme: Does the poem display a sense of rhyme?
    \item Content: How closely does the content of the poem relate to the provided acrostic words?
    \item Readability: How coherent are the vocabulary and grammar used in the poem?
\end{itemize}

Poems that fail to meet the requirements of the acrostic format often cannot form a poem. Therefore, we directly assign them zero scores in total.

\paragraph{FTime}
We evaluate the accuracy of the results, i.e., the rate of the responses exactly matching the labels.

\paragraph{Agent}
We take the inner score of the game provided by the game engine, representing the extent to which the agent has reached the final goal. This score is then divided by the maximum score to assess the performance of the agent action trace.

\paragraph{XDL}
We do not evaluate the general quality of XDL task for two reasons. Firstly, XDL dataset does not contain references, so that a similarity-based evaluation (e.g., BLEU-4) is not applicable. Secondly, XDL is a robot-controlling laboratory domain language that is too complicated and professional for a language model (e.g., GPT-4) to assess. Previous work \cite{skreta2023errors} uses preferences from experts to assess 2 models on 108 test instances, which is not feasible considering our larger number of models (19) data size (660).

\begin{figure}[htb!]
    \centering
    \includegraphics[width=0.9\linewidth]{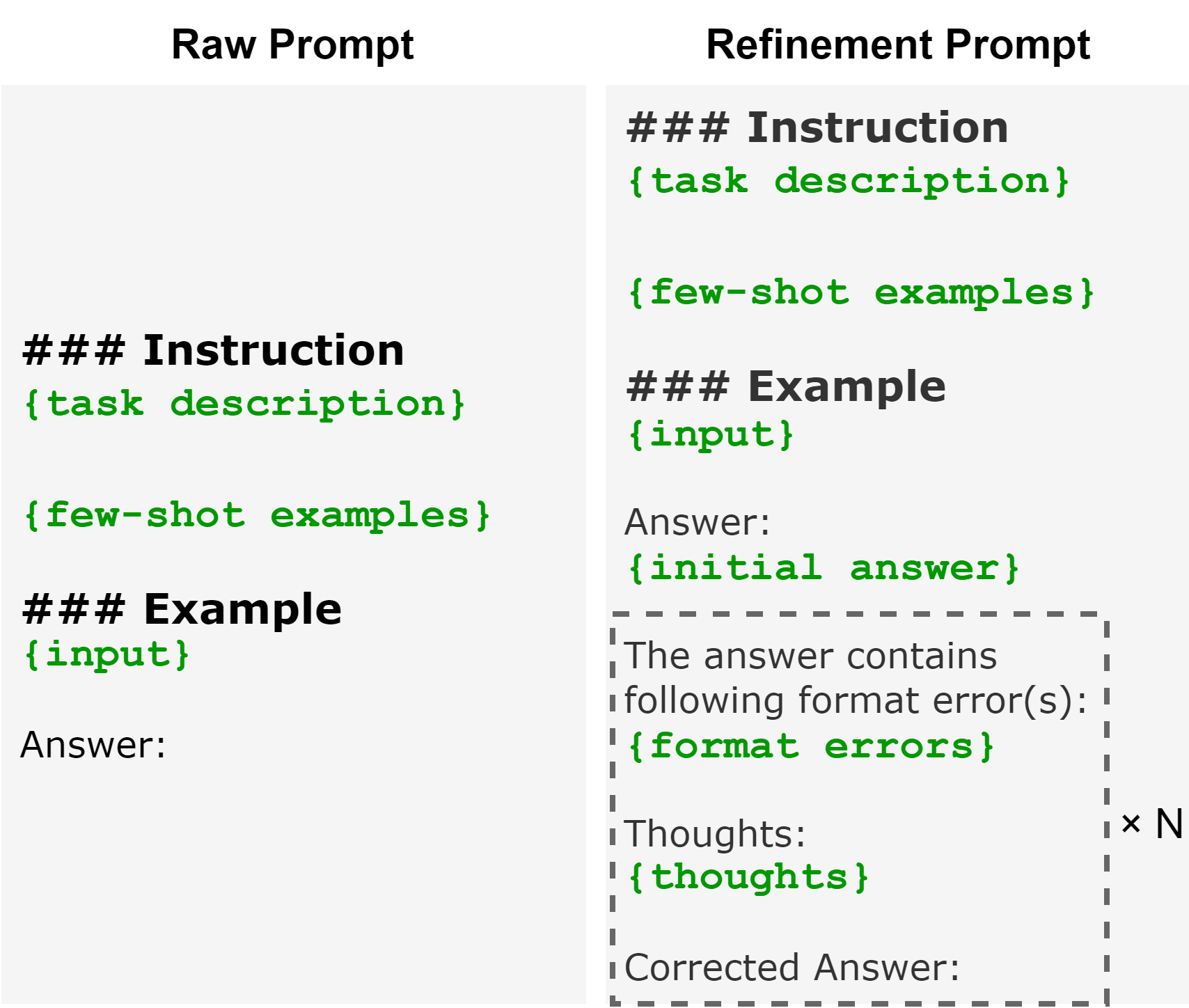}
    \caption{Prompt templates for raw generation (left, used for original models, fintuned models, and \textsc{ReFF} models) and for refinement generation (right).}
    \label{fig:prompt}
\end{figure}

\begin{figure*}[htb!]
    \centering
    \includegraphics[width=\linewidth]{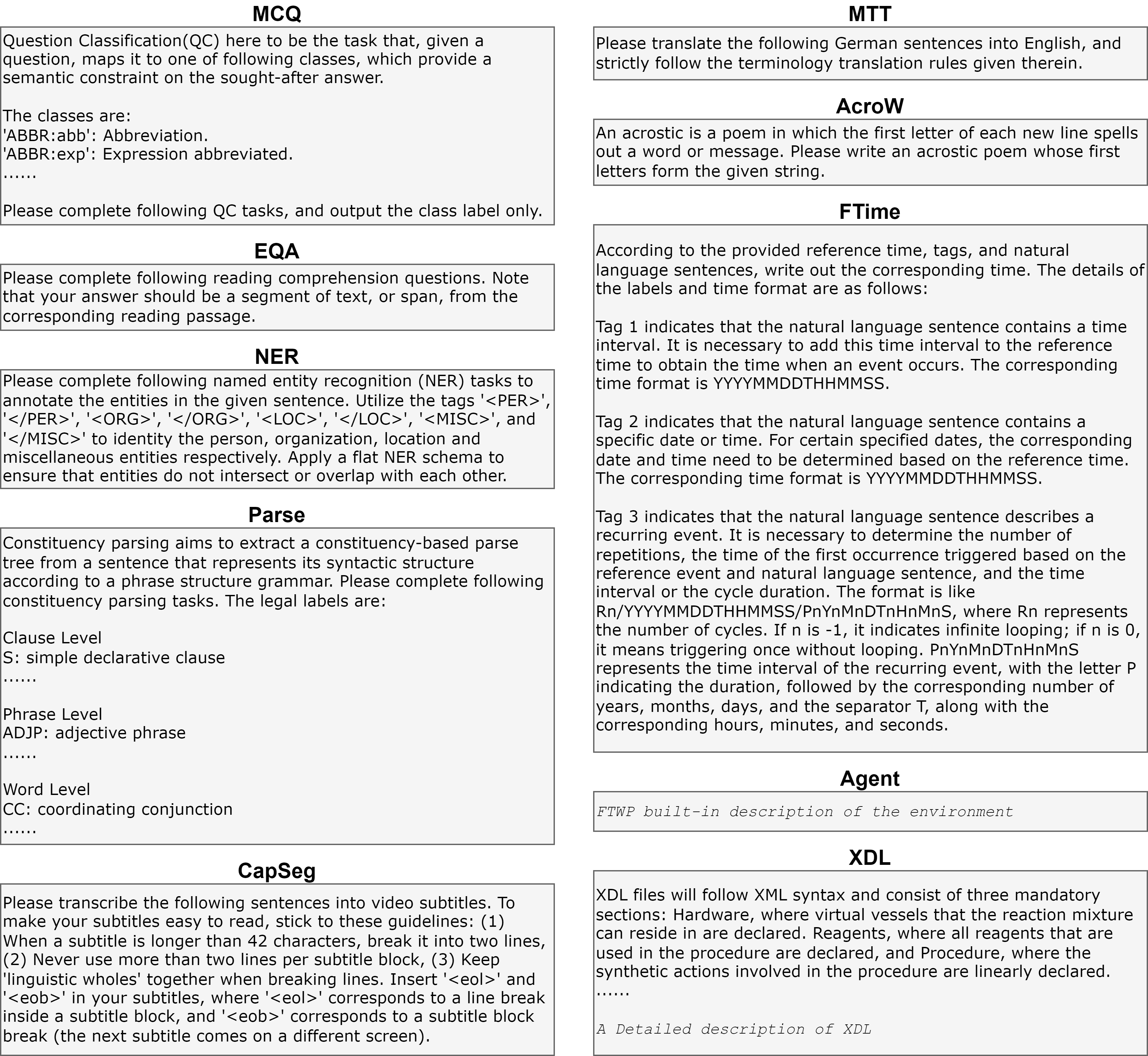}
    \caption{Task descriptions for \textsc{FormatBench}, which become a part in the prompt for each task.}
    \label{fig:task-descriptions}
\end{figure*}

\section{Math Representation of \textsc{ReFF}}
\label{app:algorithm}

Given an LLM $\mathcal{M}_\theta$, a format checker $\mathcal{F}$ (defined in Equation \ref{equ:format-checker}), and a query set $Q$, the objective function of \textsc{ReFF} to obtain an adapted model $\mathcal{M}_\phi$ is computed as follows.

\begin{equation}
    \mathbb{R}(Q; \phi) =
    \mathbb{E}_{q \in Q}\mathbb{E}_{r \sim \mathcal{M}_\phi(q)}[\mathcal{F}(q, r) - \beta \log \frac{\mathcal{M}_\phi(r \vert x)}{\mathcal{M}_\theta(r \vert x)}],
\end{equation}

where $\beta$ is a dynamically adjusting weight that regulates the KL divergence between $\mathcal{M}_\phi$ and $\mathcal{M}_\theta$ around a pre-defined value (the kl target hyper-parameter), which is set to 6 is our implementation.

\section{Prompt Designs}
\label{app:prompts}

The prompt of the vanilla approach involves combining task descriptions with few-shot examples to formulate a prompt, as is shown in Figure \ref{fig:prompt} left. The task description contains the task definition and the format specification. All the task descriptions are listed in Figure \ref{fig:task-descriptions}.

Moreover, we utilize refinement in prompt designing to improve the format faithfulness by including error messages generated by the format checker and LLM reflections. The employed prompt template is shown in Figure \ref{fig:prompt} right.

\section{Adaptation Implementation}
\label{app:implementation}

In this section, we outline the detailed implementations of (1) refinement, (2) finetuning, and (3) \textsc{ReFF}. Notably, during the entire process of LLM generation, we employ the greedy decoding strategy, where beam search is not utilized and the token with the highest probability is selected.

\subsection{Refinement}

\begin{table*}[htbp]
    \centering
    \begin{tabular}{lcccccccccc}
        \toprule
        \multirow{2}{*}{\textbf{Models}} & {\textbf{MCQ}} & {\textbf{EQA}} & {\textbf{NER}} & {\textbf{Parse}} & {\textbf{CapSeg}} & {\textbf{MTT}} & {\textbf{AcroW}} & {\textbf{FTime}} & {\textbf{Agent}} & \multirow{2}{*}{\textbf{FFR}}\\
        \cmidrule(lr){2-2} \cmidrule(lr){3-3} \cmidrule(lr){4-4} \cmidrule(lr){5-5} \cmidrule(lr){6-6} \cmidrule(lr){7-7} \cmidrule(lr){8-8} \cmidrule(lr){9-9} \cmidrule(lr){10-10}
        & \textbf{Acc} & \textbf{F1} & \textbf{F1} & \textbf{F1} & \textbf{F1} & \textbf{BLEU} & \textbf{Score} & \textbf{Acc} & \textbf{Score} \\
        \hline
        GPT-3.5      & 71.6 & 70.0 & 94.3 & 18.7 & 40.6 & 30.9 & 43.2 & 67.4 & 14.5 & 63.8 \\
        LLaMA3       & 57.6 & 67.8 & 88.3 & 0.0  & 47.3 & 32.2 & 0.7  & 40.4 & 5.7  & 54.8 \\
        Gemma        & 57.2 & 77.0 & 88.2 & 5.5  & 49.2 & 32.2 & 1.0  & 54.8 & 7.1  & 54.7 \\
        Qwen1.5      & 43.2 & 71.1 & 87.4 & 5.0  & 45.5 & 30.6 & 0.2  & 49.2 & 3.9  & 54.3 \\
        Mistral      & 59.8 & 69.3 & 86.1 & 3.6  & 47.6 & 30.7 & 4.0  & 57.3 & 4.7  & 54.2 \\
        Mistral-inst & 55.4 & 72.5 & 82.8 & 3.5  & 49.0 & 30.6 & 2.8  & 54.4 & 9.3  & 53.0 \\
        LLaMA2       & 35.8 & 61.3 & 82.8 & 0.6  & 45.7 & 29.9 & 0.1  & 42.3 & 6.3  & 51.6 \\
        LLaMA        & 25.0 & 61.8 & 81.5 & 0.2  & 11.8 & 26.9 & 0.0  & 38.9 & 2.2  & 50.1 \\
        Falcon       & 19.2 & 54.7 & 78.9 & 0.0  & 44.8 & 23.0 & 0.0  & 1.3 & 3.2  & 41.2 \\
        Falcon-inst  & 12.0 & 45.5 & 80.7 & 7.6  & 41.3 & 14.1 & 0.0  & 1.3 & 1.2  & 32.5 \\
        \bottomrule
    \end{tabular}
    \caption{General Quality on \textsc{FormatBench}, where all values are scaled by 100. The last column (FFR) stands for average format faithfulness shown in Table \ref{tab:ff-scores}.}
    \label{tab:gq-scores}
\end{table*}

The stopping criterion of the refinement process includes (1) the format compiler detects no format errors, (2) the refinement step reaches the pre-defined limit (five in our implementation), (3) the prompt exceeds the maximum length supported by the model, and (4) the model repeats one of its previous answers.

\begin{table}[htb!]
    \centering
    \begin{tabular}{lcc}
        \toprule
        \textbf{Hyper-Parameters} & \textbf{Finetuning} & \textbf{\textsc{ReFF}} \\
        \hline
        learning rate & 2e-5 & 1.41e-5 \\
        optimizer & adam & adam \\
        adam betas & (0.9, 0.999) & (0.9, 0.999) \\
        epoch & 3 & 3 \\
        batch size & 256 & 32 \\
        lr schedule & constant & constant \\
        LoRA $\alpha$ & 8 & 8 \\
        LoRA dropout & 0 & 0 \\
        LoRA rank & 16 & 16 \\
        LoRA modules & attention layers & attention layers \\
        initial kl & - & 0.05 \\
        horizon & - & 1,000 \\
        kl target & - & 6 \\
        \bottomrule
    \end{tabular}
    \caption{Hyper-parameters of finetuning and \textsc{ReFF}. The bottom three lines are specifically related to reinforcement.}
    \label{tab:hyper-parameters}
\end{table}

\subsection{Finetuning}

\paragraph{Data Processing}
We randomly sample up to 4,000 instances (both queries and references) from each task involved, and shuffle them to get the finetuning dataset.

\paragraph{Configuration}
We conduct finetuning with \texttt{trl} \cite{vonwerra2022trl} library. We employ a completion only fine-tuning, where the only the tokens of generated answers contribute to gradients. Additionally, we use a parameter-efficient tuning approach, namely Low-Rank Adaptation (LoRA) \cite{hu2021lora}. Relevant hyper-parameters are listed in Table \ref{tab:hyper-parameters}.

\paragraph{Computational Usage}
We use one NVIDIA A800 80GB GPU to conduct format tuning. The finetuning process of \textsc{ReFF}-trn-ft takes about 4 hours.

\subsection{\textsc{ReFF}}

\paragraph{Data Processing}
We randomly sample up to 4,000 instances (queries only) from each task involved, and shuffle them to get the query set.

\paragraph{Configuration}
We conduct reinforcement learning with \texttt{trl} \cite{vonwerra2022trl} library when using a parameter-efficient tuning approach, namely Low-Rank Adaptation (LoRA) \cite{hu2021lora}. Relevant hyper-parameters are listed in Table \ref{tab:hyper-parameters}.

\paragraph{Computational Usage}
We use one NVIDIA A800 80GB GPU to conduct format tuning. It takes about 200 hours to train \textsc{ReFF}-tst, and 50 hours to complete the RL process of \textsc{ReFF}-trn, and \textsc{ReFF}-trn-ft.

\section{General Quality of Original Models}
\label{app:gq}

General quality results (as defined in Appendix \ref{app:benchmark}) are shown in Table \ref{tab:gq-scores}. General quality scores and the format faithfulness rate in Table \ref{tab:ff-scores} are relatively consistent, showing that the two metrics are positively correlated under many scenarios.

\end{document}